\documentclass{article}

     \PassOptionsToPackage{numbers, compress}{natbib}

\usepackage[preprint]{neurips_2026}

\usepackage[utf8]{inputenc} 
\usepackage[T1]{fontenc}    
\usepackage{hyperref}       
\usepackage{url}            
\usepackage{booktabs}       
\usepackage{amsfonts}       
\usepackage{nicefrac}       
\usepackage{microtype}      
\usepackage{xcolor}         
\usepackage[pdftex]{graphicx} 
\usepackage{float}
\usepackage{amsmath}
\usepackage{siunitx}
\usepackage{marvosym}
\graphicspath{{images/}}
\newcommand{\Dstroke}{\text{\DJ}}

\title{It’s All Connected: Topology-Aware Structural Graph Encoding Improves Performance on Polymer Prediction}

\author{
\makebox[\textwidth][c]{%
\begin{tabular}{c}
Halil I.~Erdogan$^{1}$\,
Punith Raviswamy$^{1}$,
Nikita Agrawal$^{2}$,
Yannik K\"{o}ster$^{3,4}$, \\
Stefan Zechel$^{3,4,5,6}$,
Ulrich S.~Schubert$^{3,4,5,6}$,
Ruben Mayer$^{2}$,
Christopher Kuenneth$^{1}$\Letter, \\[0.7em]
{\small $^{1}$Faculty of Engineering Science, University of Bayreuth, Germany} \\
{\small $^{2}$Faculty of Mathematics, Physics \& Computer Science, University of Bayreuth, Germany} \\
{\small $^{3}$Laboratory of Organic and Macromolecular Chemistry (IOMC), Friedrich Schiller University Jena, Germany} \\
{\small $^{4}$Jena Center for Soft Matter (JCSM), Friedrich Schiller University Jena, Germany} \\
{\small $^{5}$Helmholtz Institute for Polymers in Energy Applications Jena (HIPOLE Jena), Germany} \\
{\small $^{6}$Helmholtz Zentrum Berlin f\"ur Materialien und Energie GmbH (HZB), Germany} \\[0.4em]
{\small \Letter\ Corresponding author: \texttt{christopher.kuenneth@uni-bayreuth.de}}
\end{tabular}%
}
}

\begin{document}

\maketitle

\begin{abstract}
Graph Neural Networks (GNNs) have achieved strong results in molecular property
prediction, but polymers present distinct challenges: 
labeled datasets are scarce and small (typically in the order of hundreds of polymers) due to the need for expensive experimentation, and complex polymer chain distributions influence polymer properties. Established practice in polymer prediction represents polymers solely by graphs of
their repeat units, discarding the chain-scale morphology that governs key properties such as the glass transition temperature ($T_\text{g}$). In this work, we propose
a principled graph construction that addresses this gap. Given a polymer's molar
mass distribution (MMD), we
sample representative chains from the Schulz-Zimm distribution and construct representative sets of large
graphs encoding chain-scale topology directly, with atoms and bonds featurized using
rich chemical descriptors. We further pretrain GNN encoders via masked
graph modeling on 100{,}000 unlabeled PSMILES strings before fine-tuning on labeled
data. On a dataset of 381 polymers (180 homopolymers and 201 copolymers), we show that
graph construction and self-supervised pretraining are jointly necessary: without
pretraining, the large graph method matches the repeat-unit baseline (\SI{28.40}{K} vs.\ \SI{28.36}{K} RMSE); with pretraining, it achieves $\SI{24.76}{K} \pm \SI{3.30}{K}$, a 5.1\%
reduction in mean error over the pretrained repeat-unit baseline ($\SI{26.08}{K} \pm \SI{4.20}{K}$,
$p < 0.001$, 30 runs). An ablation removing chemical features degrades performance
to \SI{36.65}{K}, confirming both components are essential. Results are
architecture-agnostic, holding for both GINE and GATv2 encoders.
\end{abstract}

\section{Introduction}

Machine learning (ML), particularly neural networks, has accelerated polymer discovery and design by predicting properties from chemical structures~\cite{butler2018ml, 2024_Huan}. Polymer informatics~\cite{chen2021polymergenome}, a burgeoning field at the intersection between ML and materials science, focuses on polymers ubiquitous in plastics, coatings, and electronics, targeting thermal, mechanical, and electronic properties. One prominent polymer property is glass transition temperature ($T_\text{g}$) measured in Kelvin (K): the temperature where polymers shift from rigid (glassy) to flexible (rubbery) states, dictating mechanical performance under thermal stress \cite{tao2021benchmarking}. Graph Neural Networks (GNNs) perform well in this setting by modeling polymers as graphs (atoms as nodes, and bonds as edges), with prior successes in using repeat units (monomeric building blocks) directly as the graph encoding \cite{queen2023polymer,gurnani2023polymer}. However, this approach omits the inherent complexity of polymer chain topologies, specifically the molar mass distribution (MMD). This distribution, defined by the number-average molar mass ($M_\text{n}$), weight-average molar mass ($M_\text{w}$), and dispersity ($\Dstroke$), profoundly impacts bulk macroscopic properties, including $T_\text{g}$, as e.g., postulated by the Fox-Flory relation~\cite{mark2004physical, fox1950secondorder}.

Standard GNNs are formulated as deterministic functions mapping a single graph $G$ to a target $y$, i.e., f(G) = y. In contrast, polymer systems are more naturally described as distributions over graphs induced by chain length variability. As a result, the prediction task is better viewed as learning a function over graph distributions, i.e., approximating $\mathbb{E}_{G \sim \mathcal{D}}[f(G)] \approx y$, where $\mathcal{D}$ represents the chain length distribution.  This is particularly challenging for polymers. Repeating units form theoretically endless chains of diverse lengths, making exhaustive graph encodings impractical. MMD profiles describe ensembles of chains rather than discrete structures, while branching in copolymers resists canonical graph encoding approaches. Limited experimental data further constrains representation optimization. Sophisticated augmentations beyond basic repeat-unit graphs are therefore mission-critical \cite{queen2023polymer, gurnani2023polymer}. 

We address these challenges through two complementary contributions. First, we propose representing each polymer as sets of large graphs whose chain lengths are sampled from the Schulz-Zimm molar mass distribution~\cite{rubinstein2003polymer}. Given $M_\text{n}$, $M_\text{w}$, and $\Dstroke$, we parameterize a Gamma distribution and draw multiple independent sets of chains, constructing a graph that encodes the distributional shape of chain lengths directly into the topology accessible to the GNN. Each polymer is described by a polymer SMILES (PSMILES) string~\cite{kuenneth2023polybert}, an extension of the Simplified Molecular Input Line Entry System (SMILES) representation~\cite{weininger1988smiles}, and each atom and bond in the sampled chains is encoded with rich chemical features computed via RDKit~\cite{rdkit}, including atom type, hybridization, chirality, degree, and bond order, providing the GNN with the chemical identity it needs alongside the structural context. Second, we pretrain GNN encoders via masked graph modeling on 100{,}000 unlabeled PSMILES strings, giving the encoder a structural and chemical vocabulary before any labeled $T_\text{g}$ data is seen. We show that these components are mutually necessary: without pretraining, the large graphs offer no advantage over a repeat-unit graph; with pretraining, the structural encoding yields a statistically significant improvement.

Our main contributions are:
\begin{itemize}
    \item A principled large-graph construction for polymer property prediction based on Schulz-Zimm MMD sampling~\cite{rubinstein2003polymer}, embedding chain-scale structure directly into graph topology rather than as decoder-level scalars. This is the basis for our Monte Carlo approximation of the $T_\text{g}$ expectation for polymer systems.
    \item Rich chemical node and edge feature encoding via RDKit~\cite{rdkit}, shown by ablation to be essential: a topology-only graph without chemical features yields \SI{36.65}{K} RMSE, an \SI{11.89}{K} degradation relative to the full model.
    \item An empirical analysis demonstrating that graph construction and self-supervised pretraining are jointly necessary: large graphs without self-supervised pretraining (SSL) match repeat-unit graphs, while the combination achieves $\SI{24.76}{K} \pm \SI{3.30}{K}$ RMSE vs.\ $\SI{26.08}{K} \pm \SI{4.20}{K}$ for the repeat-unit baseline ($p < 0.001$, paired $t$-test, 30 runs).
    \item Architecture-agnostic results across GINE~\cite{hu2020pretrain} and GATv2~\cite{brody2022gatv2} encoders, confirming the gain originates from the graph construction and pretraining strategy rather than model-specific inductive biases.
    \item Uncertainty quantification via Monte Carlo Dropout~\cite{gal2016dropout}, providing per-polymer predictive uncertainty estimates alongside point predictions.
\end{itemize}

\section{Background}

\paragraph{Graph Neural Networks.}
GNNs operate on graphs by iteratively aggregating information from local neighbourhoods via message passing~\cite{gilmer2017neural, xu2019gin}. At each layer $\ell$, a node $v$ updates its representation by aggregating messages from its neighbours $\mathcal{N}(v)$:
\begin{equation}
    \mathbf{h}_v^{(\ell+1)} = \mathrm{UPDATE}\left(\mathbf{h}_v^{(\ell)},\; \mathrm{AGGREGATE}\left(\{(\mathbf{h}_u^{(\ell)},\, \mathbf{e}_{uv}) : u \in \mathcal{N}(v)\}\right)\right)
\end{equation}

where $\mathbf{e}_{uv}$ are edge features. After $L$ layers, a global readout (e.g.\ mean pooling) produces a graph-level embedding, which is passed to a multi-layer perceptron (MLP) for prediction. We evaluate two message-passing variants: GINE~\cite{hu2020pretrain}, which incorporates edge features into the aggregation via a sum-based update, and GATv2~\cite{brody2022gatv2}, which uses dynamic attention coefficients to weight neighbour contributions.

\paragraph{Self-supervised pretraining.}
Self-supervised learning (SSL) on graphs trains an encoder without labels by solving auxiliary tasks defined on the graph structure itself. We use masked graph modeling, where a fraction of node and edge features are masked and the model is trained to reconstruct them~\cite{hu2020pretrain}. Pretraining on large unlabeled corpora provides the encoder a general structural vocabulary that can be transferred to downstream tasks with limited labeled data.

\paragraph{Polymer chains and molar mass distributions.}
Polymer chains are macromolecules made by polymerizing many small monomer molecules into long chains. Depending on the monomer units' functionality and reaction pathway, polymerization generates e.g., linear chains, branched, or network structures. When two or more different monomer molecule types are present, the macroscopic structure is called a copolymer with varying distributions of monomers.

Polymers or polymer materials are not just single chains but distributions of chain lengths and architectures, quantified by molar mass statistics: $M_\text{n}$ (number-average molar mass), $M_\text{w}$ (weight-average molar mass), 
and $\Dstroke = M_\text{w}/M_\text{n}$ (dispersity of chain lengths)~\cite{rubinstein2003polymer}. A $\Dstroke$ of 1 indicates perfectly uniform chains (monodisperse, rare in practice), while $\Dstroke > \num{2}$ is typical for industrial polymers, reflecting broad chain length variation within one sample. This structural polydispersity makes polymers far more complex than small molecules, as identical repeat units can yield vastly different chain configurations and bulk properties, and consequently makes it more challenging for GNNs to learn consistent representations.

Higher $M_\text{n}$ increases chain entanglement, enhancing mechanical strength but raising processing viscosity. The glass transition temperature ($T_\text{g}$), a reversible transition from rigid/glassy to flexible/rubbery states, is particularly sensitive to MMD, rising with molar mass~\cite{rubinstein2003polymer}. $\Dstroke$ broadens the distribution of $T_\text{g}$ within samples. These ensemble effects fundamentally distinguish polymers from discrete small molecules, creating unique challenges for computational property prediction.

\section{Problem Formulation}

We consider the problem of predicting a scalar polymer property (specifically the
glass transition temperature $T_\text{g} \in \mathbb{R}$) from the chemical structure of a polymer
and associated molecular descriptors. We introduce notation used throughout the paper.

\paragraph{Polymer representation.}
A polymer is described by its PSMILES string~\cite{kuenneth2023polybert}, an extension of the standard SMILES notation~\cite{weininger1988smiles} in which two dummy atoms ($[*]$) mark the polymerization attachment sites of the repeat unit. We consider two classes of polymers: homopolymers, characterized by a single repeat unit with PSMILES $s$, where the chain is formed by repeating $s$ for $n$ repetitions connected at the attachment sites ($n$ being the degree of polymerization determined by the molar mass distribution); and copolymers, characterized by two distinct repeat units with PSMILES $s_1$ and $s_2$ combined in molar fractions $\phi_1, \phi_2 \in [0,1]$ with $\phi_1 + \phi_2 = 1$. As our dataset contains only polymers with two monomer types, we consider binary copolymers corresponding to random copolymers, where repeat units are distributed randomly along the backbone according to their relative composition.

In addition to the PSMILES string, each polymer is associated with a vector of global
molecular descriptors. For homopolymers,
$\mathbf{g} = (M_\text{n},\, M_\text{w},\, \Dstroke,\, M_0) \in \mathbb{R}^4$,
where $M_\text{n}$ is the number-average molar mass, $M_\text{w}$ is the weight-average
molar mass, $\Dstroke = M_\text{w} / M_\text{n}$ is the dispersity, and
$M_0$ is the repeat-unit molar mass. For copolymers, the descriptor is extended
to $\mathbf{g} = (M_\text{n},\, M_\text{w},\, \Dstroke,\, M_0,\, \phi_1,\, \phi_2) \in \mathbb{R}^6$.
For homopolymers, we set $(\phi_1, \phi_2) = (1, 0)$, so all polymers share a
uniform 6-dimensional input to the decoder.

\paragraph{Graph representation.}
We represent a polymer as a graph
$\mathcal{G} = \bigl(\mathcal{V},\, \mathcal{E},\, \mathbf{X},\, \mathbf{E}\bigr)$,
where $\mathcal{V}$ is the set of nodes (atoms),
$\mathcal{E} \subseteq \mathcal{V} \times \mathcal{V}$ is the set of edges (bonds), treated as undirected,
$\mathbf{X} \in \mathbb{R}^{|\mathcal{V}| \times d_v}$ is the node feature matrix
with $d_v$-dimensional atom features per node, and
$\mathbf{E} \in \mathbb{R}^{|\mathcal{E}| \times d_e}$ is the edge feature matrix
with $d_e$-dimensional bond features per edge. The specific choice of node and edge
feature encodings, together with the graph topology (i.e., which atoms and bonds
are included), constitute the \emph{graph construction} and are the central focus
of this work.

\paragraph{Task.}
Given a dataset
$\mathcal{D} = \{(\mathcal{G}_i,\, \mathbf{g}_i,\, T_{\mathrm{g},i})\}_{i=1}^{N}$
of $N$ labeled polymers, the goal is to learn a function
\begin{equation}
    f:\; (\mathcal{G},\, \mathbf{g}) \;\longrightarrow\; \hat{T}_\text{g} \in \mathbb{R}
    \label{eq:task}
\end{equation}
that minimizes the mean squared error
$\frac{1}{N}\sum_{i=1}^{N}\bigl(\hat{T}_{\text{g},i} - T_{\text{g},i}\bigr)^2$
over the dataset.

\paragraph{Key question.}
The central challenge is not the choice of GNN architecture for $f$, but rather the
\emph{graph construction}: given a PSMILES string, how should $\mathcal{G}$ be
defined such that the resulting representation is highly informative for property
prediction? Unlike small molecules, polymers are macromolecules with no single
well-defined molecular graph. The repeat unit captures only local chemical identity,
while chain-scale structures, such as chain topology distributions and monomer sequences, critically
determine macroscopic properties such as $T_\text{g}$. This tension between local
chemistry and chain-scale physics and chemistry is the core difficulty in polymer graph
construction, and it is the problem we address in this work.

\section{Why Standard Graph Constructions Fall Short}
\label{sec:limitations}

\paragraph{Scalar descriptors alone are insufficient for GNN embeddings.}
The dominant approach in GNN-based polymer property prediction is to represent a
polymer by the graph of its repeat unit~\cite{kuenneth2023polybert,
chen2021polymergenome}. In this construction, nodes correspond to atoms within a
single repeat unit and edges correspond to bonds, with polymerization
attachment sites removed. Global molecular descriptors such as $M_\text{n}$, $M_\text{w}$, and
$\Dstroke$ are typically appended as scalar features at the graph readout stage.
This approach is computationally convenient but introduces a fundamental
representational bottleneck. Appending $M_\text{n}$, $M_\text{w}$, and $\Dstroke$ as scalars to the decoder can shift
the final prediction globally, but it cannot alter the GNN embeddings produced by
message passing. The message-passing layers operate solely on the repeat-unit
graph and therefore produce identical node embeddings for all polymers that share
the same repeat unit, regardless of how different their chain lengths or molecular
weight distributions are. Since $T_\text{g}$ and other polymer properties are governed by chain-scale physics (chain
length, entanglement density, free volume) that emerge only from the collective
arrangement of many repeat units~\cite{rubinstein2003polymer}, this means the
GNN is asked to predict a chain-scale property from a graph that contains no
chain-scale information. The scalar descriptors are incorporated too late in the pipeline
to compensate. They reach the decoder after the structural embedding is already
fixed. Recent work~\cite{kimmig2026} addresses this by using coarse-grained monomer-level graphs with kinetic Monte Carlo-generated chain ensembles; our approach differs in operating at the atom level with full bond-feature encoding and self-supervised pretraining.

\paragraph{Empirical evidence.}
We confirm this limitation experimentally. Despite both models receiving identical
scalar descriptors ($M_\text{n}$, $M_\text{w}$, $\Dstroke$, $M_0$), the repeat-unit
construction underperforms our proposed approach when both are combined with
self-supervised pretraining, which encodes the molar mass distribution
directly into the graph topology. The improvement is
statistically significant ($p < 0.001$, paired $t$-test) and holds across two
distinct GNN architectures. Full experimental details and results are reported in
Section~\ref{sec:results}.

\section{Proposed Graph Construction}
\label{sec:method}

\begin{figure}[t]
    \centering
    \includegraphics[width=0.9\linewidth]{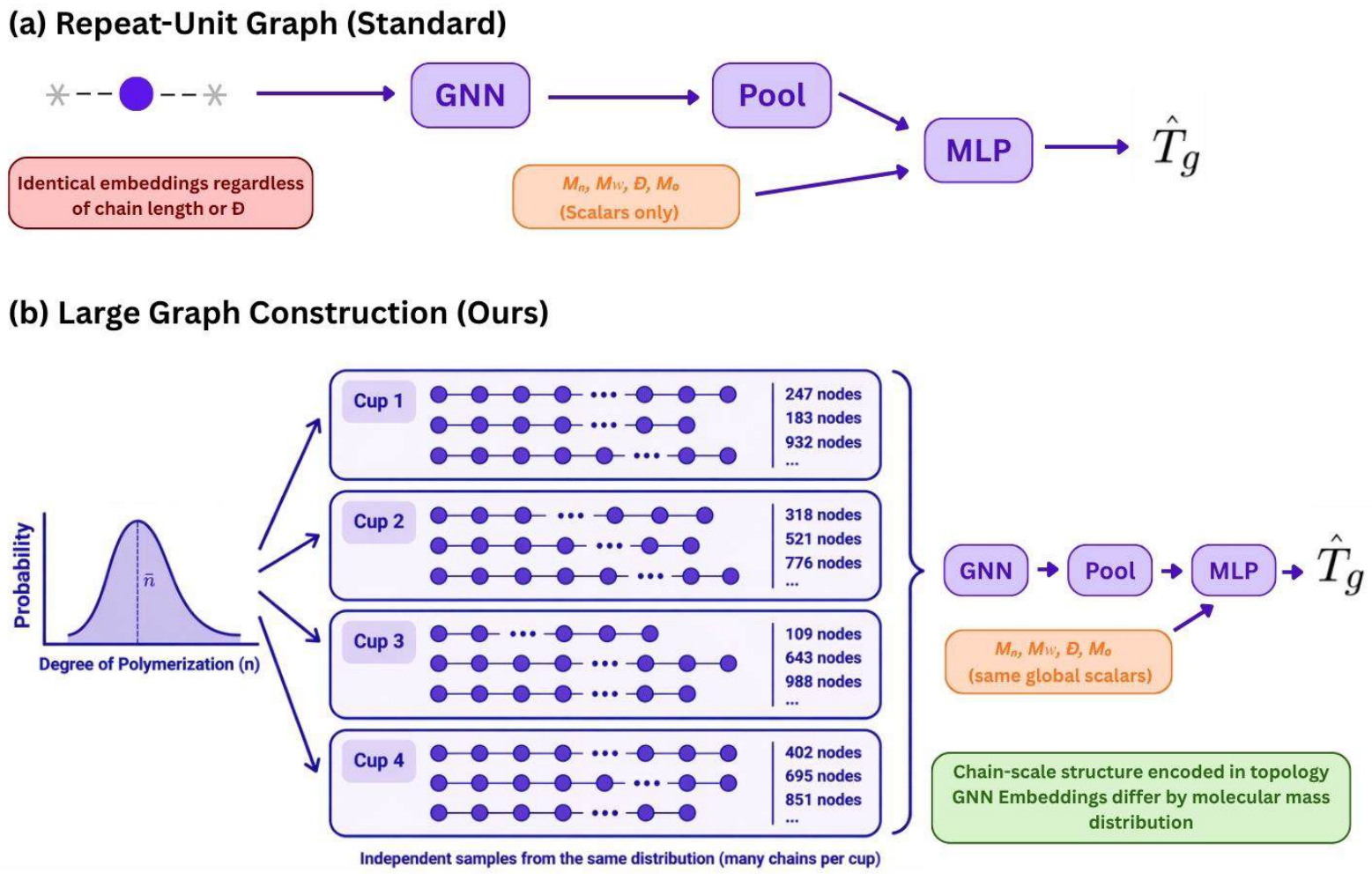}
    \caption{Comparison of graph construction strategies. \textbf{(a)} The standard
    repeat-unit graph produces identical node embeddings for identical repeat-unit graphs, regardless of
    chain length or $\Dstroke$; molar mass information enters only as global scalars.
    \textbf{(b)} Our large-graph construction samples chain lengths from the
    Schulz-Zimm (Gamma) distribution, embedding chain-scale structure directly
    into the graph topology accessible to message-passing layers.}
    \label{fig:concept}
\end{figure}

We propose representing a polymer not as a single repeat-unit graph but as an
ensemble of full polymer chains whose lengths are sampled from the molar
mass distribution of the sample. This embeds chain-scale structure, physically
determined by $M_\text{n}$, $M_\text{w}$, and $\Dstroke$, directly into the graph topology,
making it accessible to the GNN's message-passing layers rather than available
only as global scalars at the decoder (Figure~\ref{fig:concept}).

\paragraph{Step 1: Parameterizing the molar mass distribution.}
Real polymer samples follow a broad distribution of chain lengths described by
the Schulz-Zimm distribution~\cite{rubinstein2003polymer}, which can be expressed as a Gamma distribution. Given $M_\text{n}$, $M_\text{w}$, and monomer mass $M_0$, we
compute the number-average degree of polymerization $\mathrm{DP}_\text{n} = M_\text{n} / M_0$
and the dispersity $\Dstroke = M_\text{w} / M_\text{n}$, and parameterize
$\mathrm{Gamma}(k,\,\theta)$ as:
\begin{equation}
    k = \frac{1}{\Dstroke - 1}, \qquad \theta = \frac{\mathrm{DP}_\text{n}}{k}.
    \label{eq:gamma}
\end{equation}
To avoid computationally intractable chains, we rescale $\theta$ by a factor
$\min\!\left(1,\, \mathrm{DP}_{\max} / F^{-1}_{\mathrm{Gamma}}(\num{0.99};\, k, \theta)\right)$,
which caps the 99th-percentile DP at $\mathrm{DP}_{\max} = \num{1000}$ while preserving
the shape parameter $k$ and therefore the qualitative form of the distribution.

\paragraph{Step 2: Sampling chains (cups).}
For each polymer we generate $C = \num{4}$ independent stratified samples from the parameterized distribution, which we call \emph{cups} (independent sampled graph instances). Each cup contains approximately 30
chains whose DPs are drawn via stratified inverse-CDF sampling over 8 equal-probability
quantile bins of the distribution. This can be viewed as a stratified Monte Carlo
sampling scheme, providing balanced coverage across both short- and long-chain regimes. Multiple cups per polymer serve as structural
replicas, capturing the stochastic variability of the molar mass
distribution.

\paragraph{Step 3: Building chain graphs and feature encoding.}
Each chain of DP $n$ is constructed by extracting the repeat unit from the
PSMILES string~\cite{kuenneth2023polybert} (removing the two dummy attachment atoms),
repeating it $n$ times and connecting adjacent units via explicit bonds at the attachment
sites. All chains within a cup are concatenated into a single (disconnected) graph $\mathcal{G}$, where each chain forms a separate connected component. A GNN with global mean pooling then produces a single polymer embedding from the full multi-chain graph, integrating information across all sampled chain lengths. Each atom is encoded as a $d_v = 31$-dimensional feature vector comprising:
atom type (12 classes), hybridization (4), chirality (3), degree (number of bonds, 7),
aromaticity, ring membership, hydrogen count, formal charge, and normalized
atomic mass. Each bond is encoded as a $d_e = 14$-dimensional vector comprising:
bond type (5 classes), conjugation, ring membership, and stereochemistry (7).
All features are computed with the cheminformatics tool RDKit~\cite{rdkit}.

\paragraph{Copolymers.}
For copolymers with two repeat units $s_1$, $s_2$ and molar fractions
$\phi_1$, $\phi_2$, the effective monomer mass is
$M_0^{\mathrm{eff}} = \phi_1 M_{0,1} + \phi_2 M_{0,2}$, which is used in
place of $M_0$ in Equation~\ref{eq:gamma}. Each chain position is assigned
repeat unit $s_1$ or $s_2$ independently with probabilities $\phi_1$ and
$\phi_2$, yielding a statistical copolymer sequence consistent with the
experimentally measured composition.

\paragraph{Global features.}
In addition to the graph, we pass the scalar descriptors
$\mathbf{g} = (M_\text{n},\, M_\text{w},\, \Dstroke,\, M_0)$ directly to the decoder after
global mean pooling. These scalars complement the structural encoding: The graph
topology encodes the distributional shape of chain lengths as structural patterns
that the GNN can learn through pretraining, while the scalars provide the decoder
with precise numerical values of the same quantities.

\section{Experimental Setup}
\label{sec:setup}

\begin{figure}[t]
    \centering
    \includegraphics[width=\linewidth]{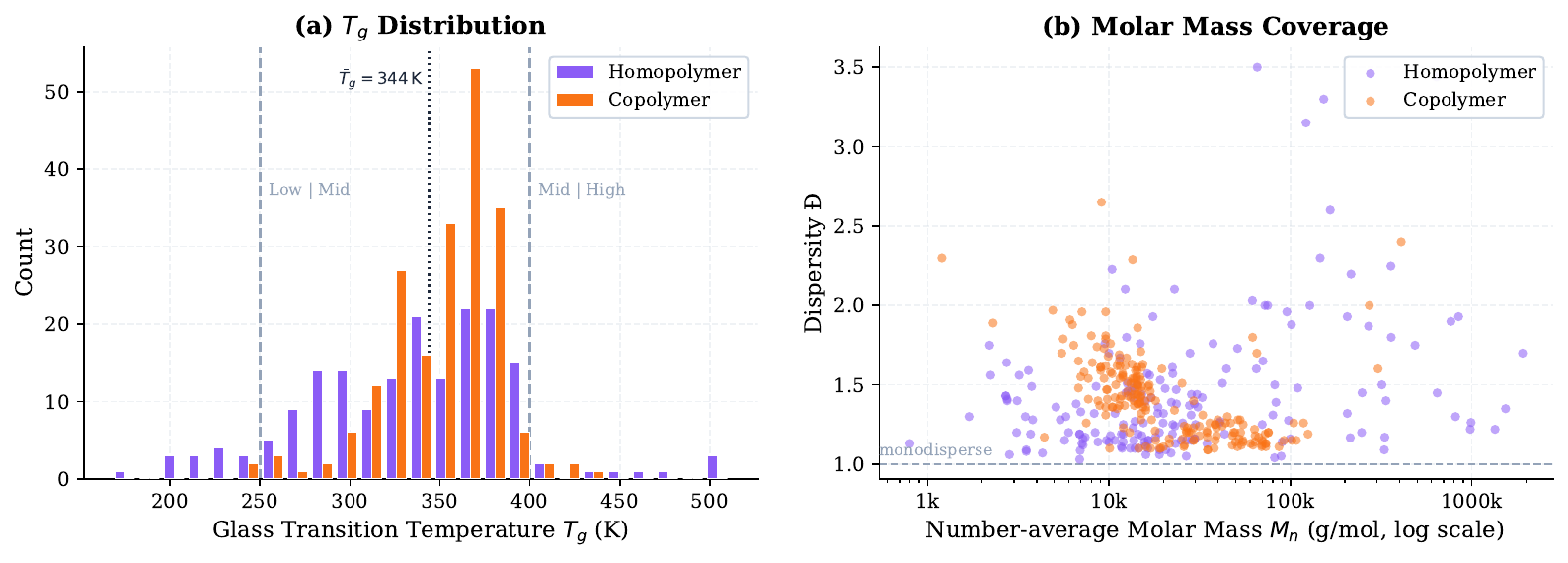}
    \caption{Dataset statistics for the 381-polymer labeled set.
    \textbf{Left:} Distribution of experimentally measured $T_\text{g}$ values
    for homopolymers (purple) and copolymers (orange), with stratum
    boundaries at 250\,K and \SI{400}{K} indicated by dashed lines.
    \textbf{Right:} Number-average molar mass $M_\text{n}$ vs.\
    dispersity $\Dstroke$ (log scale), illustrating the wide range of
    molar mass distributions present in the dataset.}
    \label{fig:dataset}
\end{figure}

\paragraph{Labeled dataset.}
Polymer records, each comprising repeat-unit chemical identity, molar mass
descriptors ($M_\text{n}$, $M_\text{w}$, $\Dstroke$), and an experimentally
measured $T_\text{g}$, were compiled from two polymer reference
handbooks~\cite{wypych2012handbook, brandrup1999polymer}, the PHA copolymer
dataset of \citet{jiang2020pha}, and experimental data contributed by
co-authors at Friedrich Schiller University Jena (Y.~K\"{o}ster, S.~Zechel, U.~S.~Schubert). The labeled dataset comprises
381 polymer samples (180 homopolymers and 201 copolymers) with experimentally
measured glass transition temperatures spanning $T_\text{g} \in [\SI{173}{K},\, \SI{506}{K}]$
(mean $\SI{344}{K} \pm \SI{46}{K}$). Molar mass descriptors cover a wide range:
$M_\text{n} \in [\SI{800}{\gram\per\mol},\, \SI{1.905882e6}{\gram\per\mol}]$ (median \num{16500}), $\Dstroke \in [\num{1.0},\, \num{3.5}]$
(mean \num{1.4}), and $M_0 \in [\SI{28}{\gram\per\mol},\, \SI{462}{\gram\per\mol}]$. The broad span of $M_\text{n}$ and $\Dstroke$
values motivates the use of a distribution-aware graph construction rather than
fixed scalar descriptors. Each polymer is represented by $C=4$ independent sampling instances (“cups”), yielding 1,524 graphs in total. Dataset statistics are visualized in
Figure~\ref{fig:dataset}. For self-supervised pretraining we sample 100{,}000 PSMILES~\cite{kuenneth2023polybert} strings from
the polyOne dataset~\cite{zenodo7124187}, which contains hypothetical
polymers generated by fragment-based combinatorial composition. All strings are chemically valid as ensured by the dataset construction but largely unsynthesized (i.e., without experimental $T_\text{g}$ labels), providing a diverse pretraining
signal without requiring experimental labels.

\paragraph{Models.}
We evaluate two GNN architectures to demonstrate that our graph construction generalizes across architectures. Both share the same encoder design: 3 message-passing layers with
hidden dimension 32, GraphNorm~\cite{cai2021graphnorm}, GELU activations, residual connections, and dropout ($p=0.1$). The two variants are:
\begin{itemize}
    \item \textbf{GINE}~\cite{hu2020pretrain}: Graph Isomorphism Network with edge
    features (primary model).
    \item \textbf{GATv2}~\cite{brody2022gatv2}: Graph Attention Network v2 with
    edge features (generalization check).
\end{itemize}
The shared decoder applies global mean pooling over node embeddings, concatenates
the 6-dimensional global feature vector $\mathbf{g}$, and passes the result
through an MLP ($38 \rightarrow 16 \rightarrow 1$, where 38 = 32-dimensional graph embedding + 6 global features) with GELU activation and
dropout ($p=0.45$).

\paragraph{Self-supervised pretraining.}
Both encoders are pretrained via masked graph modeling on the 100{,}000 unlabeled corpus. We randomly mask 15\% of nodes and 15\% of edges independently per graph and train
separate prediction heads to recover masked atom types (11-class cross-entropy)
and masked bond types (5-class cross-entropy). Pretraining uses AdamW~\cite{loshchilov2019adamw}
($\mathrm{lr}=5\times10^{-4}$, $\lambda=10^{-4}$), batch size 64, cosine
annealing LR, and early stopping with patience 10 over 65 epochs. The pretrained
encoder weights are then transferred to the downstream model.

\paragraph{Fine-tuning.}
Downstream $T_\text{g}$ prediction is trained with AdamW~\cite{loshchilov2019adamw} ($\mathrm{lr}=5\times10^{-5}$,
$\lambda=3\times10^{-4}$) and ReduceLROnPlateau (factor 0.3, patience 30,
$\mathrm{lr}_{\min}=10^{-6}$). We use Huber loss with weighted sampling to
balance the $T_\text{g}$ distribution. Training runs for up to 85 epochs with early
stopping (patience 30) after a minimum of 15 warm-up epochs. Batch size is 2
graphs; since each graph encodes an entire cup of polymer chains (tens to hundreds
of chains, up to $\mathrm{DP}_{\max}=\num{1000}$ repeat units per chain), the effective
number of atoms per update step is large. Target $T_\text{g}$ values are transformed
to an approximately normal distribution using a quantile transformer and
inverse-transformed before computing RMSE, so all reported errors are in Kelvin.

\paragraph{Evaluation protocol.}
We report test RMSE in Kelvin (K). Polymers are split using stratified 5-fold
cross-validation, with folds balanced across three $T_\text{g}$ strata: Low
($T_\text{g} < 250$\,K), Mid ($250 \leq T_\text{g} < 400$\,K), and High ($T_\text{g} \geq 400$\,K).
All cups of a given polymer are kept within the same fold to prevent data
leakage. Cup-level predictions are averaged to produce one polymer-level
prediction before computing RMSE. We repeat each experiment with 6 random seeds
(30 runs total) and report mean $\pm$ standard deviation. The same seeds and
fold assignments are used for all compared methods. All experiments are run on a
single NVIDIA A100-SXM4-40GB GPU. Graph construction for the labeled dataset takes \SI{228}{s} in total (\SI{134}{s} for 201 copolymers, \SI{94}{s} for 180 homopolymers). SSL corpus graph creation (100K trimers) takes \SI{42}{min}; SSL encoder training is performed once and incurs a one-time computational cost. Fine-tuning (5-fold CV) takes approximately \SI{128}{min} per seed, for a total of approximately \num{12.8}~GPU-hours across all 30 runs. At inference, the SSL model predicts $T_\text{g}$ for one polymer in $\SI{5.66}{ms} \pm \SI{0.80}{ms}$ (mean $\pm$ std over 5 folds, measured on the test set of approximately 75 polymers per fold).

\section{Results}
\label{sec:results}

\paragraph{Pretraining unlocks the structural benefit.}
Table~\ref{tab:main_results} summarizes test RMSE across all settings.
Without SSL pretraining, the large-graph and repeat-unit constructions perform
equivalently (\SI{28.40}{K} vs.\ \SI{28.36}{K}, not significant), showing that a
larger graph alone does not help when training from scratch on 381 labeled
polymers. The structural richness of the Schulz-Zimm-sampled chains must first
be learned through pretraining before it can be exploited at fine-tuning time.

With SSL pretraining, the picture changes substantially. Our large-graph
construction achieves $\SI{24.76}{K} \pm \SI{3.30}{K}$ across 30 evaluation runs
(6 seeds $\times$ 5 folds), compared to $\SI{26.08}{K} \pm \SI{4.20}{K}$ for the
repeat-unit baseline under the same protocol, a 5.1\% reduction in mean
error with strictly lower variance. A paired $t$-test on the 30 matched runs
yields $t = 3.90$ ($p < 0.001$, $df = 29$), and the large-graph construction
outperforms the repeat-unit baseline in 24 of 30 matched runs. The repeat-unit graph cannot encode
chain-scale structure in its node embeddings regardless of how it is trained;
SSL pretraining on diverse unlabeled chains gives the large-graph encoder the structural patterns it needs to exploit the molar mass distribution at
inference time.

\paragraph{Architecture generalization.}
To verify that the benefit is not specific to GINE, we evaluate the same
construction with a GATv2~\cite{brody2022gatv2} encoder. GATv2 achieves
$\SI{24.73}{K} \pm \SI{2.75}{K}$ (30 runs), nearly identical to GINE (\SI{24.76}{K}) and
well below the repeat-unit baseline. Both models share identical graph
construction, pretraining protocol, and decoder; only the message-passing
mechanism differs. The consistent improvement across two architecturally
distinct GNNs confirms that the gain originates from the graph construction
and pretraining strategy, not from any architecture-specific inductive bias.

\paragraph{Ablation: the role of chemical features.}
To isolate the contribution of atom and bond feature encoding, we evaluate a
degenerate variant in which all node and edge features are replaced by constant
all-ones vectors, retaining the Schulz-Zimm chain topology but stripping all
chemical identity. This topology-only model achieves $\SI{36.65}{K} \pm \SI{4.38}{K}$
(5 folds, 1 seed, base training). The \SI{11.89}{K} gap relative to the full model
is unambiguous: chain topology alone is far from sufficient, and rich chemical
feature encoding is essential. Structural and chemical information are
complementary: The topology determines which atoms interact across chain-scale
distances, while atom and bond features determine what those interactions mean.
We also ablate the number of independently sampled chains per polymer (cups) used
during training and inference. Increasing cups from 1 to 3 reduces mean RMSE
from \SI{26.06}{K} to \SI{24.48}{K}, a 6.1\% improvement, as additional samples
better cover the molar mass distribution. Beyond 3 cups the mean error plateaus;
we select 4 cups as it recovers fold-to-fold variance relative to 3 cups
(std: \SI{3.30}{K} vs.\ \SI{3.66}{K}) at no additional accuracy cost.

\begin{figure}[t]
    \centering
    \includegraphics[width=\linewidth]{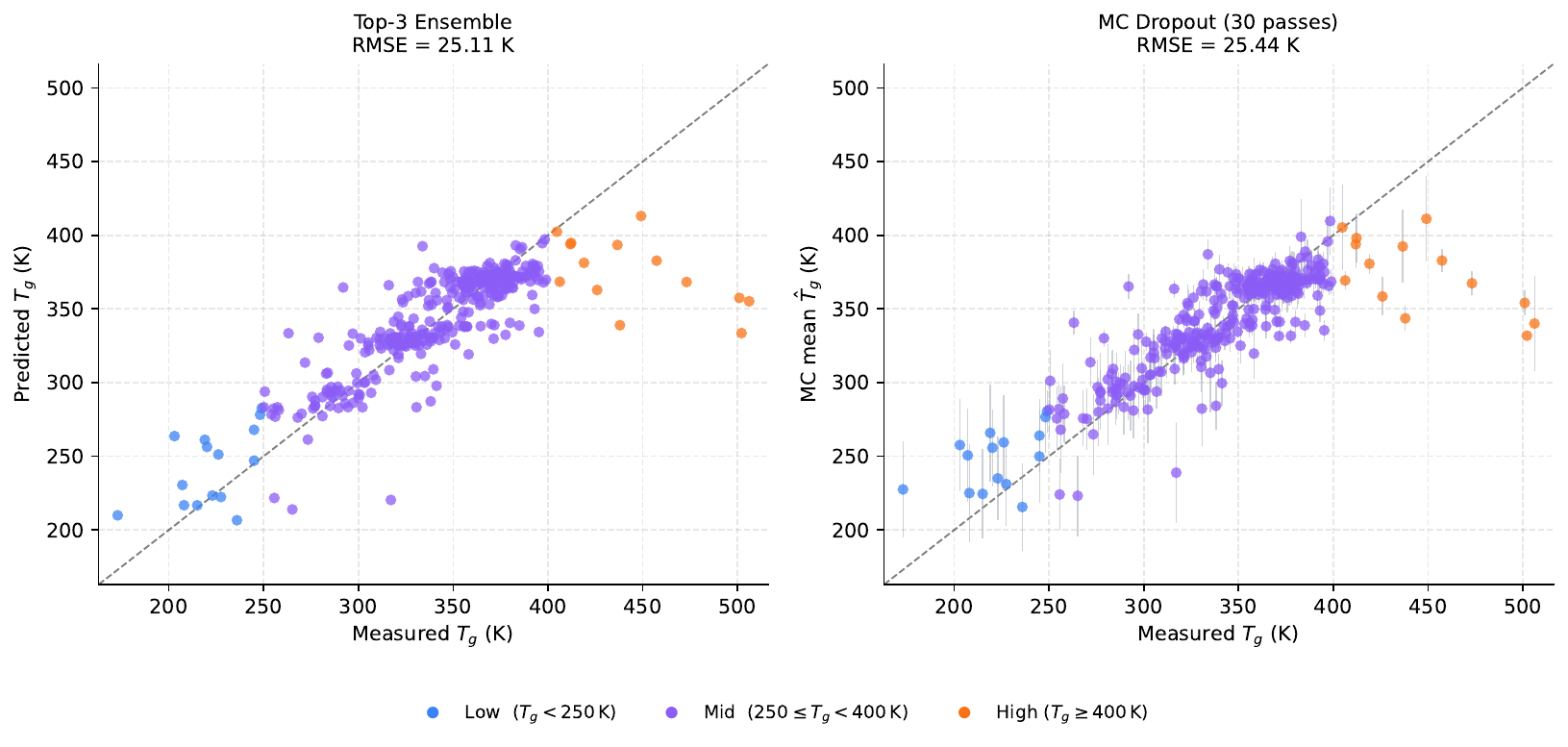}
    \caption{Predicted vs.\ measured $T_\text{g}$ for all 381 polymers (out-of-fold predictions from one representative seed). \textbf{Left:} top-3 checkpoint ensemble predictions.
    \textbf{Right:} Monte Carlo Dropout mean with $\pm$1 std uncertainty bars
    (30 stochastic forward passes). Points are colored by $T_\text{g}$ stratum:
    Low ($T_g < 250$\,K, blue), Mid ($250 \leq T_g < 400$\,K, purple),
    High ($T_g \geq 400$\,K, orange).}
    \label{fig:scatter_mc}
\end{figure}

\paragraph{Uncertainty estimation.}
We apply Monte Carlo Dropout~\cite{gal2016dropout} to the SSL model with
30 stochastic forward passes per polymer (dropout active during inference).
Each polymer receives a predictive mean $\bar{T}_\text{g}$ and sample standard
deviation $\sigma_{\mathrm{MC}}$ as a proxy for epistemic uncertainty.
Figure~\ref{fig:scatter_mc} shows predicted versus measured $T_\text{g}$ values for all
381 polymers (out-of-fold predictions), with MC uncertainty as vertical
error bars. Uncertainty is elevated at both tails of the $T_\text{g}$ distribution: in
the high-$T_\text{g}$ regime ($T_\text{g} \geq 400$\,K) and in the
low-$T_\text{g}$ regime ($T_\text{g} < 250$\,K), both of which have fewer
training examples than the mid-$T_\text{g}$ stratum.

\paragraph{Summary.}
Table~\ref{tab:main_results} summarizes all methods. Three findings are
robust: (i) the large-graph construction unlocks significant improvement
over repeat-unit graphs, but only in combination with SSL pretraining;
(ii) chemical atom and bond features are indispensable alongside chain
topology; (iii) the benefit is architecture-agnostic, holding for both
GINE and GATv2 encoders.

\begin{table}[t]
\centering
\caption{Test RMSE (K) on the 381-polymer dataset. All results are mean $\pm$ pooled std
over 30 runs (6 seeds $\times$ 5 folds). $^\dagger$Topology-only uses 1 seed, 5 folds. Lower is better.}
\label{tab:main_results}
\begin{tabular}{llcr}
\toprule
\textbf{Graph construction} & \textbf{Encoder} & \textbf{SSL} & \textbf{RMSE (K)} \\
\midrule
Topology-only (no chem.\ feats.)$^\dagger$  & GINE  & No  & $\SI{36.65}{K} \pm \SI{4.38}{K}$ \\
Topology-only (no chem.\ feats.)$^\dagger$  & GATv2 & No  & $\SI{46.76}{K} \pm \SI{2.04}{K}$ \\
Repeat-unit + global scalars                & GINE  & No  & $\SI{28.36}{K} \pm \SI{4.53}{K}$ \\
Repeat-unit + global scalars                & GATv2 & No  & $\SI{28.47}{K} \pm \SI{4.73}{K}$ \\
Large graph (ours) + globals                & GINE  & No  & $\SI{28.40}{K} \pm \SI{4.34}{K}$ \\
Large graph (ours) + globals                & GATv2 & No  & $\SI{26.96}{K} \pm \SI{4.15}{K}$ \\
\midrule
Repeat-unit + global scalars                & GINE  & Yes & $\SI{26.08}{K} \pm \SI{4.20}{K}$ \\
Repeat-unit + global scalars                & GATv2 & Yes & $\SI{25.52}{K} \pm \SI{3.26}{K}$ \\
\textbf{Large graph (ours) + globals}       & \textbf{GINE} & Yes & $\mathbf{\SI{24.76}{K} \pm \SI{3.30}{K}}$ \\
\textbf{Large graph (ours) + globals}       & \textbf{GATv2} & Yes & $\mathbf{\SI{24.73}{K} \pm \SI{2.75}{K}}$ \\
\bottomrule
\end{tabular}
\end{table}

\section{Conclusion}
\label{sec:conclusion}

We investigated the problem of graph construction for polymer property prediction, using
glass transition temperature ($T_\text{g}$) as the target property. Standard practice represents
a polymer by the graph of its repeat unit alone, discarding the chain-scale structure
that governs $T_\text{g}$ through chain entanglement and molar mass effects. Instead, we proposed
to construct a large graph by sampling chains from the Schulz-Zimm molecular
weight distribution parameterized by ($M_\text{n}$, $M_\text{w}$, $\Dstroke$), and to encode
each atom and bond with rich chemical features via RDKit.

Our central empirical finding is that graph construction and self-supervised pretraining
are jointly necessary given the scale of our labeled dataset. Without SSL, large graphs
and repeat-unit graphs perform identically (\SI{28.40}{K} vs.\ \SI{28.36}{K}), showing that
the richer chain-scale topology offers no benefit when trained from scratch on only
381 labeled polymers. This likely reflects the limited scale of the labeled dataset rather than a fundamental limitation of the construction: a GNN trained on a few hundred samples cannot learn how chain-length patterns relate to the target property from supervision alone. With SSL pretraining on
100{,}000 unlabeled PSMILES, the encoder acquires a chain-scale structural
vocabulary before any $T_\text{g}$ labels are seen, and the large-graph construction
then achieves $\SI{24.76}{K} \pm \SI{3.30}{K}$ RMSE, a statistically significant improvement
over the repeat-unit baseline ($\SI{26.08}{K} \pm \SI{4.20}{K}$, $p < 0.001$, paired $t$-test,
30 runs). Whether the large-graph construction would yield direct gains without SSL
on substantially larger labeled datasets is an open question we were unable to answer
with the data available, and is an important direction for future work.

An ablation stripping chemical features from the large graph degrades performance
to \SI{36.65}{K}, confirming that chain topology and chemical identity are both
indispensable. The benefit is architecture-agnostic: GINE and GATv2 achieve
nearly identical results (\SI{24.76}{K} and \SI{24.73}{K}), making a model-specific explanation less likely.

Our construction may underperform in settings where multi-scale structure does
not govern the target property, where the distributional descriptors
($M_\text{n}$, $\Dstroke$) are unavailable or unreliable, or where the
distribution is very narrow so that sampled chains are nearly identical.

Beyond the settings explored here, the full polymer-graph construction enables several
capabilities over repeat-unit-only graphs that the present dataset does not yet exercise. First, end groups (the
terminal chemical units that cap each chain) can be incorporated as additional nodes,
which is especially consequential for low-$M_\text{n}$ polymers where the end-group
mass fraction is non-negligible; this representation also extends naturally to short
biological macromolecules such as DNA, RNA, and proteins, where chain termini carry
distinct chemistry. Second, branched, star, and hyperbranched polymer architectures
are naturally encodable by varying the sampled chain topology, whereas repeat-unit
graphs can only approximate such structures through ad hoc modifications. Third,
tacticity (the stereochemical arrangement of side groups along the backbone, whether
isotactic, syndiotactic, or atactic) is a sequence-level property requiring multiple
consecutive repeat units to be present in the graph; once stereochemical sequence information is available in the input representation,
for example through BigSMILES-style macromolecular notation~\cite{lin2019bigsmiles},
the full-chain representation can encode it without architectural changes, with
relevance beyond $T_\text{g}$ to crystallinity and phase behavior~\cite{chang2010tacticity}. Fourth, copolymer microstructure,
whether block (AAABBB) or random (ABABBA), is directly encodable as graph topology,
allowing models to distinguish architectures that exhibit markedly different
$T_\text{g}$ signatures~\cite{kim2006gradient}; our current dataset contains only
statistically random copolymers, making this a natural direction for future data
collection and modeling.

These results suggest a broader principle: in data-scarce materials settings where
the relevant structural descriptor is a distribution rather than a single canonical
structure, encoding representative samples from that distribution directly into
graph topology, combined with self-supervised pretraining on unlabeled structural
data, is a viable and effective strategy. Extending this to other polymer properties,
other material classes, or larger labeled datasets remains an important direction
for future work.

\section*{Acknowledgements}
This work was funded by the European Research Council (ERC) under the 
European Union's Horizon Europe programme (grant agreement No.\ 101220388, 
project genPI). The authors also thank the Federal Ministry for Economic 
Affairs and Energy (BMWE, project Digi-RoM, funding number: 03EE2072D) 
for financial support.

\section*{Usage of Generative AI}
Generative AI was used solely for language editing (grammar, spelling, and clarity). It did not generate scientific content, perform calculations, or alter the authors' interpretations, and all revisions were reviewed and approved by the authors.

\section*{Conflicts of Interest}
The authors declare no conflicts of interest.

\section*{Data Availability}
The datasets used in this study are described in Section~\ref{sec:setup}. The code and model weights will be made publicly available in the Kuenneth Group GitHub repository (\url{https://github.com/kuennethgroup}) upon publication of this work.

\bibliographystyle{unsrtnat}
\bibliography{references}

\newpage
\section{Appendix}
\label{sec:appendix}

\subsection*{Model Sensitivity Analysis: What the GNN Learns About Dispersity and Chain Length}
\label{app:sensitivity}

\begin{figure}[h]
    \centering
    \includegraphics[width=\linewidth]{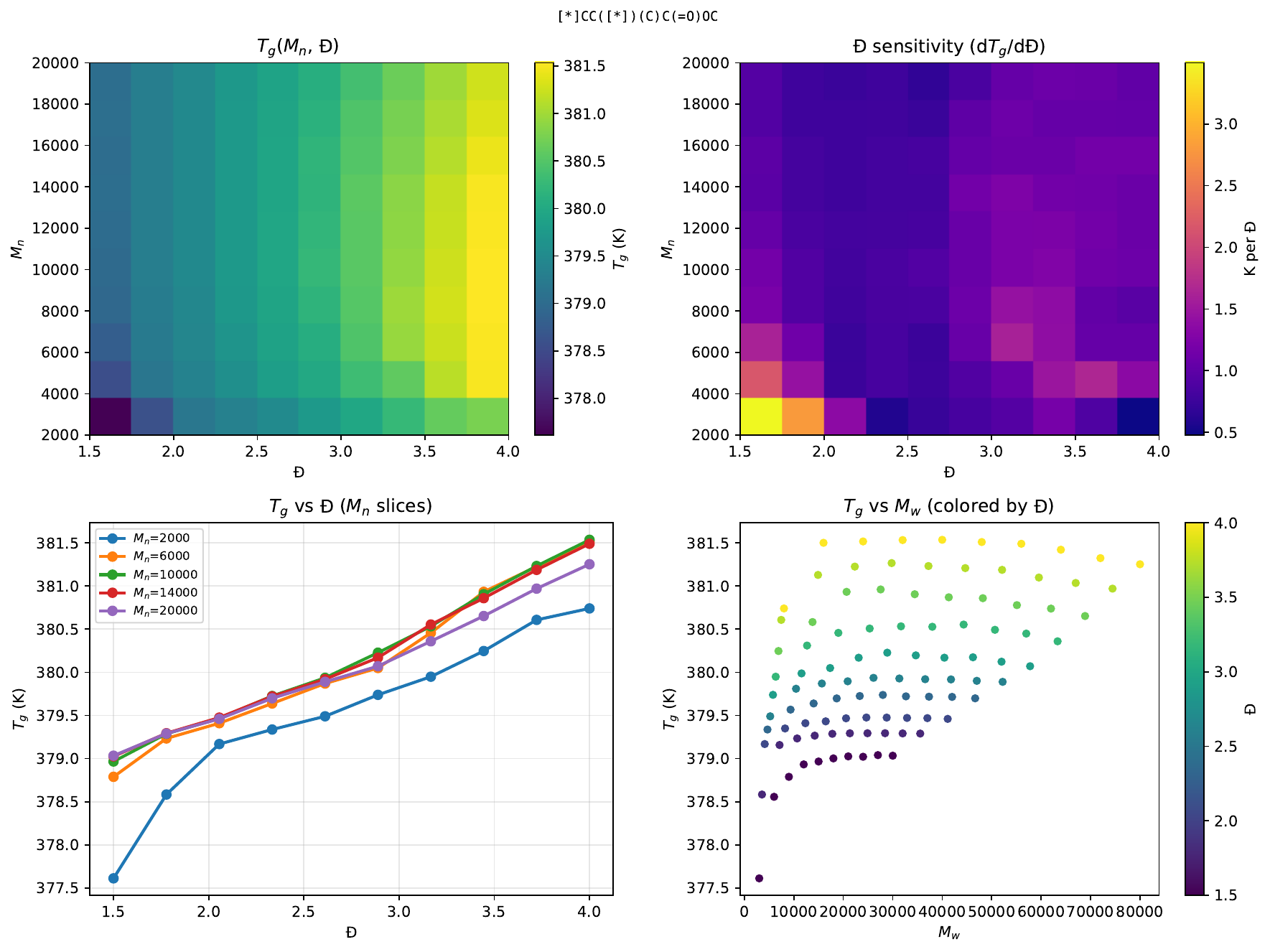}
    \caption{Sensitivity sweep for PSMILES \texttt{[*]CC([*])(C)C(=O)OC}.
    Note the crossing $M_\text{n}$ slices at low $\Dstroke$ (bottom-left): higher $M_\text{n}$
    yields lower $\hat{T}_\text{g}$ in the low-dispersity regime, a non-monotonic interaction
    invisible to additive models. Sensitivity magnitude is low throughout ($<$3\,K per $\Dstroke$
    unit) with a patchy, non-uniform spatial structure.}
    \label{fig:sweep_poly2}
\end{figure}

\begin{figure}[h]
    \centering
    \includegraphics[width=\linewidth]{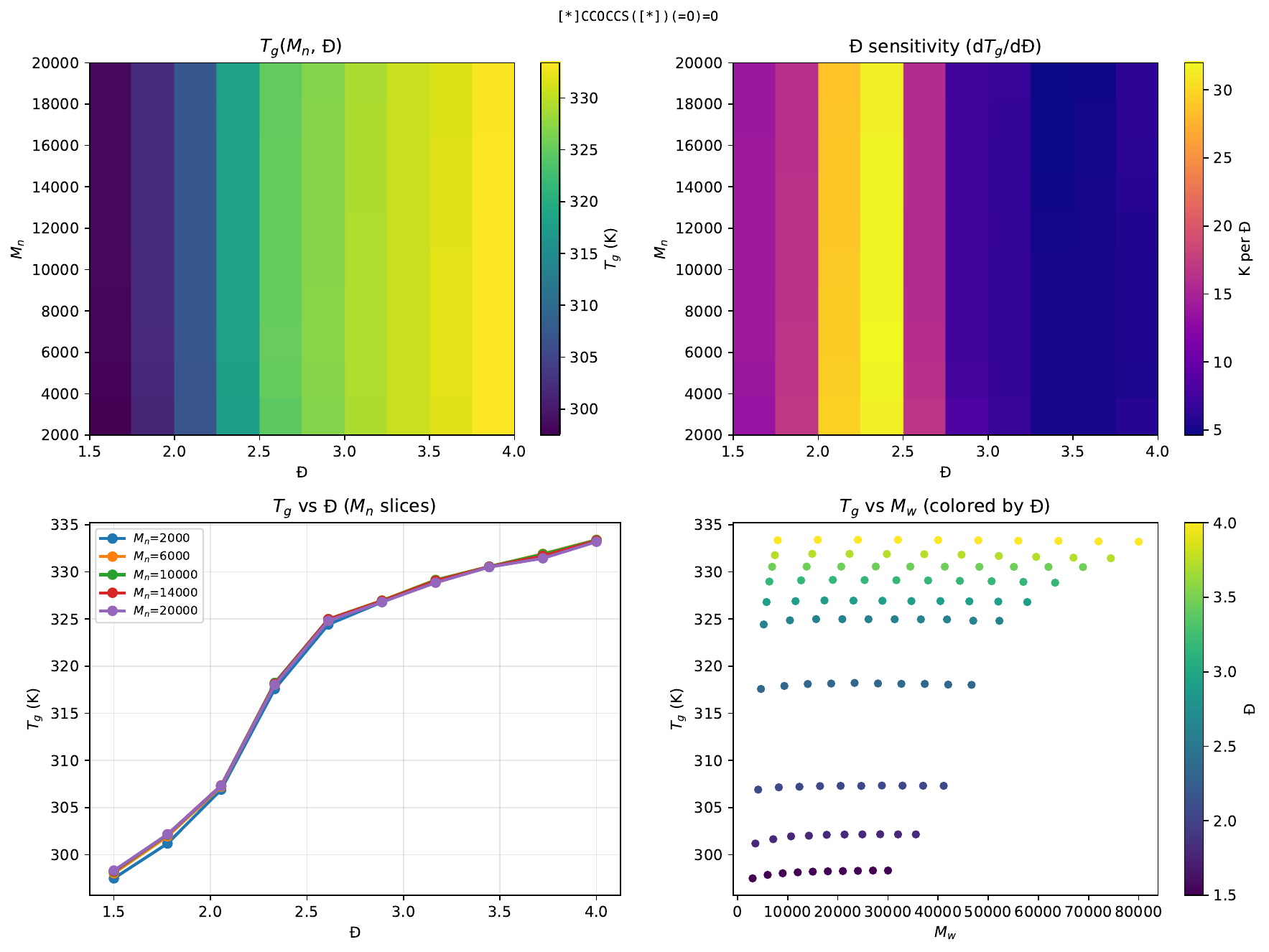}
    \caption{Sensitivity sweep for PSMILES \texttt{[*]CCOCCS([*])(=O)=O}.
    $\hat{T}_\text{g}$ spans ${\sim}37$\,K across the grid and sensitivity reaches up to
    30\,K per $\Dstroke$ unit at low $\Dstroke$, an order of magnitude larger than in
    Figure~\ref{fig:sweep_poly2}. The concave $\hat{T}_\text{g}$--$\Dstroke$ curves
    (bottom-left) show diminishing returns as $\Dstroke$ increases.}
    \label{fig:sweep_poly4}
\end{figure}

A natural question is whether the model has merely learned a simple monotonic rule
(e.g., ``higher $\Dstroke$ always raises $T_\text{g}$ by a fixed amount'') or whether it has
internalized a more physically nuanced relationship between the molecular weight distribution
and the $T_\text{g}$.
To probe this, we construct a controlled sweep: for five homopolymers selected from the training
distribution, we generate a $10{\times}10$ grid spanning $M_\text{n} \in [2{,}000,\,20{,}000]$
and $\Dstroke \in [1.5,\,4.0]$, giving 500 synthetic entries with no observed $T_\text{g}$.
We run inference with a single trained fold checkpoint and compute the finite-difference
sensitivity $\partial \hat{T}_\text{g} / \partial \Dstroke$ at each grid point.
We highlight two contrasting polymers in Figures~\ref{fig:sweep_poly2} and~\ref{fig:sweep_poly4}.

\paragraph{Non-linearity and saturation.}
The bottom-left panels show $\hat{T}_\text{g}$ as a function of $\Dstroke$ at fixed
$M_\text{n}$ slices. In both polymers the response is concave: the largest gains in
$\hat{T}_\text{g}$ occur at low $\Dstroke$, and the curve flattens as $\Dstroke$ increases.
A linear model would produce parallel straight lines with equal slope everywhere; the model
instead predicts diminishing returns, consistent with the physical intuition that once the
molecular weight distribution is broad enough, further polydispersity has a progressively
smaller effect on chain entanglement and hence on $T_\text{g}$.

\paragraph{$M_\text{n}$--$\Dstroke$ interaction and polymer-specific magnitude.}
The sensitivity heatmaps (top-right panels) reveal that
$\partial \hat{T}_\text{g} / \partial \Dstroke$ is not constant across the
$(M_\text{n},\, \Dstroke)$ plane.
For the polymer in Figure~\ref{fig:sweep_poly4}, sensitivity reaches up to ${\sim}30$\,K per $\Dstroke$
unit, an order of magnitude larger than for the polymer in Figure~\ref{fig:sweep_poly2} ($<$3\,K per $\Dstroke$ unit throughout).
This contrast shows the model has learned chemistry-dependent sensitivity rather than
a universal shift applied uniformly to all repeat units.

\paragraph{Non-monotonic $M_\text{n}$ effect.}
In Figure~\ref{fig:sweep_poly2}, the $M_\text{n}$ slices in the line plot cross at low
$\Dstroke$: higher $M_\text{n}$ yields a lower predicted $T_\text{g}$ in that regime.
This non-monotonic interaction, which would be invisible to any model that treats $M_\text{n}$
and $\Dstroke$ as independent additive inputs, suggests the GNN has captured a coupling between
chain stiffness, steric bulk, and chain-length effects that produces qualitatively different
behavior depending on the dispersity regime.

Taken together, these results indicate that the model has not collapsed to a simple linear or
separable function of the molecular weight descriptors. The graph-based representation combined
with global descriptor conditioning allows the model to express interactions between local
chemistry and chain-scale statistics that are consistent with polymer chemistry.

\subsection*{Atom and Bond Feature Encoding}

Tables~\ref{tab:atom_features} and~\ref{tab:bond_features} list all features used
to construct the node feature matrix $\mathbf{X} \in \mathbb{R}^{|\mathcal{V}| \times 31}$
and edge feature matrix $\mathbf{E} \in \mathbb{R}^{|\mathcal{E}| \times 14}$.
All features are computed with RDKit~\cite{rdkit}. One-hot vectors append an
``other'' bin for out-of-vocabulary values.

\begin{table}[h]
\centering
\caption{Atom (node) features. Total dimension $d_v = 31$.}
\label{tab:atom_features}
\begin{tabular}{llr}
\toprule
\textbf{Feature} & \textbf{Encoding} & \textbf{Dim} \\
\midrule
Atom type       & One-hot: [$*$, B, C, N, O, F, P, S, Cl, Br, I, other] & 12 \\
Hybridization   & One-hot: [SP, SP$^2$, SP$^3$, other]                  & 4  \\
Chirality       & One-hot: [CW, CCW, unspecified]                        & 3  \\
Degree          & One-hot: [0, 1, 2, 3, 4, 5, other]                    & 7  \\
Is aromatic     & Binary scalar                                          & 1  \\
Is in ring      & Binary scalar                                          & 1  \\
Total \# of Hs  & Integer scalar                                         & 1  \\
Formal charge   & Integer scalar                                         & 1  \\
Atomic mass     & Scalar (divided by 100)                                & 1  \\
\midrule
\textbf{Total}  &                                                        & \textbf{31} \\
\bottomrule
\end{tabular}
\end{table}

\begin{table}[h]
\centering
\caption{Bond (edge) features. Total dimension $d_e = 14$.}
\label{tab:bond_features}
\begin{tabular}{llr}
\toprule
\textbf{Feature} & \textbf{Encoding} & \textbf{Dim} \\
\midrule
Bond type       & One-hot: [single, double, triple, aromatic, other]              & 5 \\
Stereo          & One-hot: [none, Z, E, cis, trans, any, other]                   & 7 \\
Is conjugated   & Binary scalar                                                   & 1 \\
Is in ring      & Binary scalar                                                   & 1 \\
\midrule
\textbf{Total}  &                                                                 & \textbf{14} \\
\bottomrule
\end{tabular}
\end{table}

\clearpage

\end{document}